\documentclass[conference]{IEEEtran}
\IEEEoverridecommandlockouts
\usepackage{cite}
\usepackage{amsmath,amssymb,amsfonts}
\usepackage{algorithm}
\usepackage{algpseudocode}
\usepackage{algorithmicx}
\usepackage{graphicx}
\usepackage{textcomp}
\usepackage{xcolor}
\def\BibTeX{{\rm B\kern-.05em{\sc i\kern-.025em b}\kern-.08em
    T\kern-.1667em\lower.7ex\hbox{E}\kern-.125emX}}

\makeatletter
\newcommand{\linebreakand}{%
  \end{@IEEEauthorhalign}
  \hfill\mbox{}\par
  \mbox{}\hfill\begin{@IEEEauthorhalign}
}
\makeatother

\IEEEaftertitletext{\vspace{-1\baselineskip}}

\begin{document}

\title{Online Action Representation using Change Detection and Symbolic Programming\\
}

\author{\IEEEauthorblockN{Vishnu S Nair}
\IEEEauthorblockA{\textit{Department of Electrical Engineering} \\
\textit{Indian Institute of Technology Madras}\\
hiwishnu@gmail.com}
\and
\IEEEauthorblockN{Sneha Sree}
\IEEEauthorblockA{\textit{Department of Electrical Engineering} \\
\textit{Indian Institute of Technology Madras}\\
ee21s049@smail.iitm.ac.in}
\and
\IEEEauthorblockN{Jayaraj Joseph}
\IEEEauthorblockA{\textit{Department of Electrical Engineering} \\
\textit{Indian Institute of Technology Madras}\\
jayaraj@ee.iitm.ac.in}

\linebreakand

\IEEEauthorblockN{Mohanasankar Sivaprakasam}
\IEEEauthorblockA{\textit{Department of Electrical Engineering} \\
\textit{Indian Institute of Technology Madras}\\
mohan@ee.iitm.ac.in
}
}



\maketitle

\begin{abstract}

This paper addresses the critical need for online action representation, which is essential for various applications like rehabilitation, surveillance, etc. 
The task can be defined as representation of actions as soon as they happen in a streaming video without access to video frames in the future.
Most of the existing methods use predefined window sizes for video segments, which is a restrictive assumption on the dynamics. The proposed method employs a change detection algorithm to automatically segment action sequences, which form meaningful sub-actions and subsequently fit symbolic generative motion programs to the clipped segments. 
We determine the start time and end time of segments using change detection followed by a piece-wise linear fit algorithm on joint angle and bone length sequences. Domain-specific symbolic primitives are fit to pose keypoint trajectories of those extracted segments in order to obtain a higher level semantic representation. 
Since this representation is part-based, it is complementary to the compositional nature of human actions, i.e., a complex activity can be broken down into elementary sub-actions. 
We show the effectiveness of this representation in the downstream task of class agnostic repetition detection. 
We propose a repetition counting algorithm based on consecutive similarity matching of primitives, which can do online repetition counting. 
We also compare the results with a similar but offline repetition counting algorithm. 
The results of the experiments demonstrate that, despite operating online, the proposed method performs better or on par with the existing method.

\end{abstract}


\section{Introduction}

Action representation is important for many tasks like surveillance, human-computer interface, rehabilitation monitoring, sports analysis, behavior analysis, autonomous navigation, etc.
In most of the existing methods, processing is done on fixed-sized segments of an incoming video, and an output will be produced for each segment. Some methods do segmentation to get meaningful windows as a pre-processing step, but they are mostly offline methods, i.e., they require the whole video to find the segments.
But for tasks like rehabilitation feedback, surveillance, etc., it's crucial that the methods work online. Setting a fixed window size is not optimal, as different motions have different dynamics. Certain approaches use image-by-image methods, but they don't capture the temporal dynamics of actions. We identify this gap in the existing works and propose an online action representation method, where we use change detection to segment out action sequences and then use primitive fitting to represent the motion segments at a higher level. 
The segmentation works online as it uses an online change detection algorithm. Thus, we get meaningful segments of actions as soon as they happen in the streaming video without needing future video frames.
Also, since representation is derived over time periods, they capture temporal information.


\begin{figure}[htpb]
\centering
\includegraphics[width=\columnwidth]{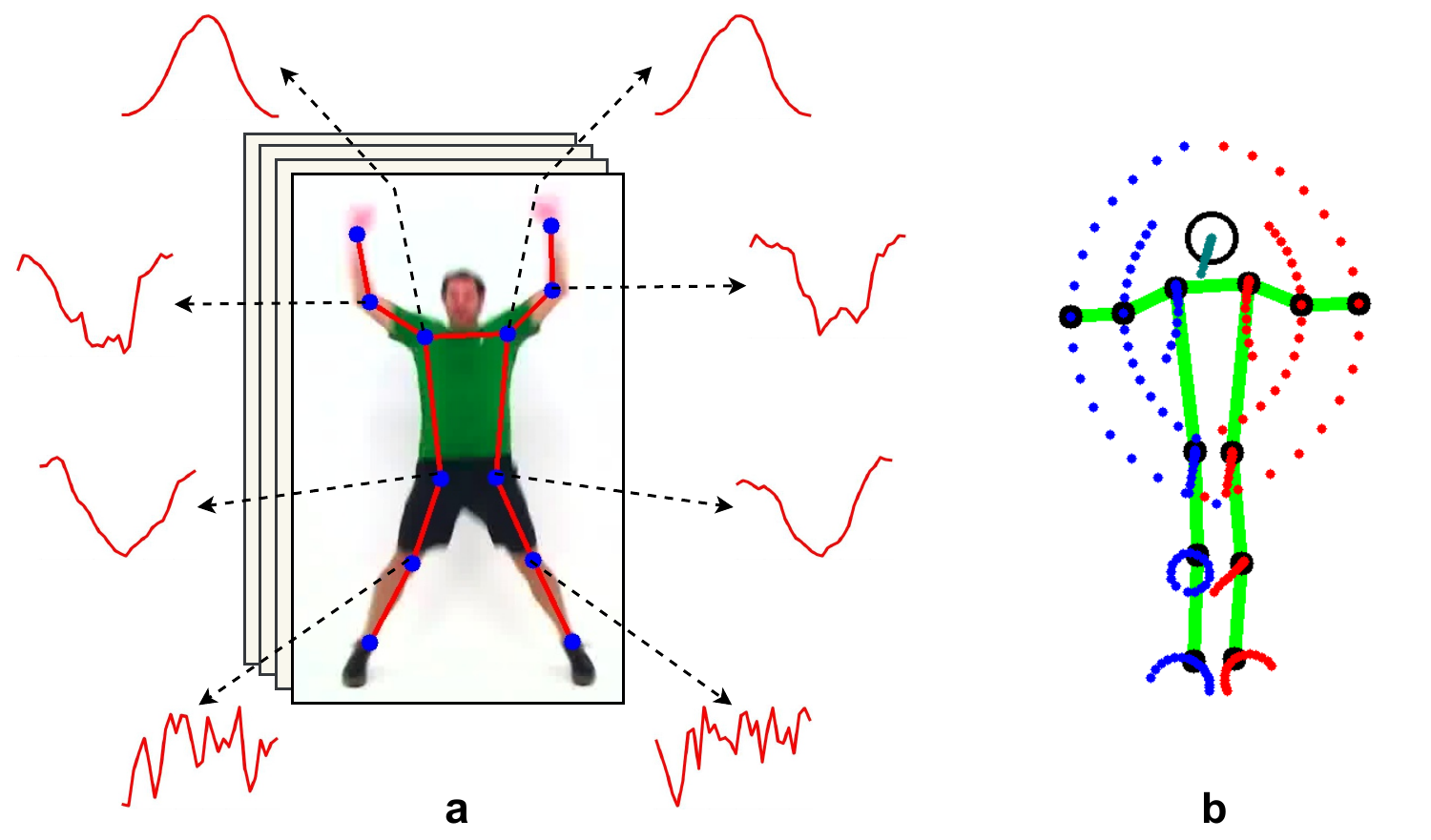} 
\caption{(a) Joint angle sequences over one instance of jumping jack \\(b) Illustration of primitives fit to each keypoint}
\label{fig_angles_prims}
\end{figure}

As an introduction, we intend to provide an intuition to the method. The motivation for this representation comes from the idea that an activity(e.g., walking) is composed of elementary sub-actions like moving left leg forward, moving right leg forward, etc. Also, these sub-actions can be detected from the increasing and decreasing changes in the joint angles that occur along with the action. So, we can extract the regions of such sub-actions and then make higher-level representations of them. Such a representation will be both discriminative and generalizable because of its compositionality. In order to segment the sub-actions, we first identify the regions of change. Next, we fit each segment to symbolic primitives to create more semantic representations. 
Symbolic programming uses domain knowledge to predefine generative symbols and fit the data optimally to those symbols. 
Here, we fit the trajectories to simple shapes like lines, circles, etc. 
These symbols are generative as they can be used for synthesis as well. 
So, with such fits, we will be able to state something like: ``the arm is rotated in this time segment." This method can work sequentially on an incoming video by the choice of an online change detection algorithm. Fig. \ref{fig_block_diagram} shows the complete pipeline of our proposed method.


We explain each step in the method in more detail. 
Later on, we use repetition detection as an example downstream task to show the effectiveness of the representation. We also compare the results of our method with the existing offline counterpart. 

\section{Related Works}

Existing works that attempt to do rehabilitation and exercise monitoring include \cite{Sabater2021, Zhao2023, Fieraru2021}, but they are not online. 
In \cite{Dittakavi2022}, an image-based approach is used that gives feedback on each frame. As it is image-based, it fails to capture the temporal information. Being data-driven makes these methods less feasible in specific domains where data is sparse, such as rehabilitation.


In \cite{Kulal2021}, a motion programming idea of fitting primitive motions to action segments is introduced. That method uses a dynamic programming approach to automatically segment the videos based on an optimal error of fitting. But this method only works offline. We improve on this by replacing the segmentation part with a change detection algorithm and then fitting a single primitive to the segmented part.  

Change detection is the process of identifying meaningful changes in a time series over time. They can be classified into online and offline algorithms. Though we use a classical change detection algorithm, it can be replaced with any online change detection method.   

Loop/repetition detection is the task of counting the number of repetitions of approximately the same action in an input video sequence. Some of the existing works on loop detection are \cite{Levy2015, Dwibedi2020}. In \cite{Levy2015}, online loop detection is done, but it uses windows of a predetermined fixed size.

\section{Methods}

\begin{figure}[htpb]  
\centering
\includegraphics[width=6.9cm]{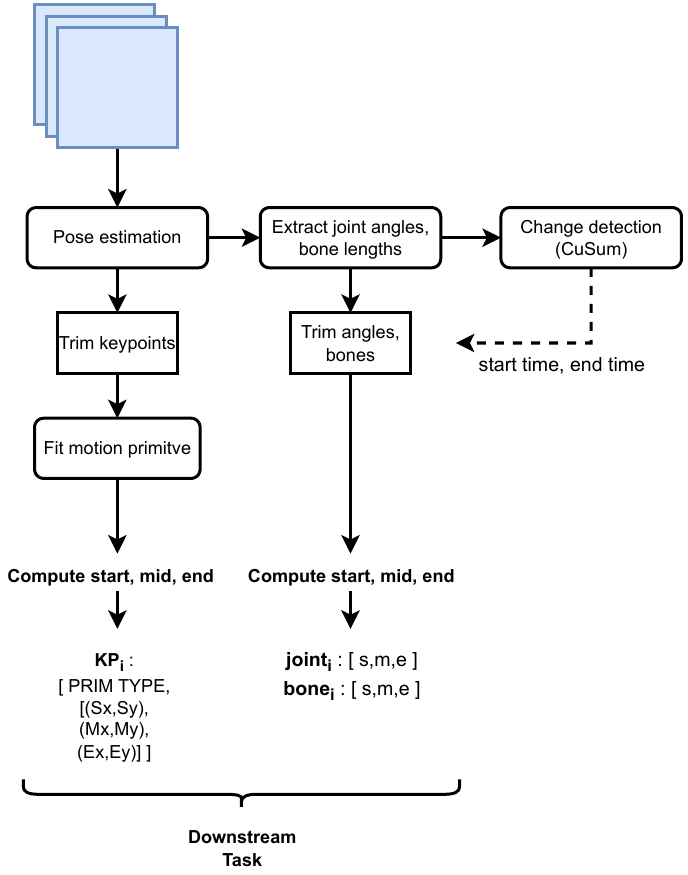}
\caption{Action representation pipeline}
\label{fig_block_diagram}
\end{figure}

Here we provide a comprehensive overview of the entire pipeline, followed by detailed explanations of relevant parts. The primary objective of our work is to make the task of action recognition online. The input is an incoming sequence of frames. 

First, we employ an off-the-shelf pose estimator to extract the keypoints from the frames. Here we use the mediapipe pose estimator\cite{mediapipe}. We calculate the angles at several joints and also the length of bones using vectors constructed from the keypoints. Joint angles refer to angles subtended at body joints, such as the shoulder angle and the elbow angle, while bone lengths refer to the lengths of body parts between two joints, such as the calf length and the torso length. Consequently, we have several sequences of joint angles and bone lengths as illustrated in Fig.\ref{fig_angles_prims}(a). These values are computed frame by frame, and hence they are online incoming time series signals. We use the CuSum change detection algorithm\cite{Page1954} to detect change in each of these signals. Once a change is triggered in any one of the signals, that particular signal is utilized to determine the starting and ending times for temporal segmentation. While CuSum provides the time of change and the starting time of change, it does not provide the ending time. Therefore, a piece-wise line-fitting method is devised to estimate the ending time after waiting for a few more frames from the point of change. 
The number of extra frames to take is decided based on a heuristic. We observe that the clip between the starting and the ending time of the detected change forms a meaningful action segment. The keypoint sequences, the joint angle sequences, and the bone length sequences are trimmed according to those time points. In order to obtain a higher-level representation of the action, we use a motion primitive fitting method. This method adopts a symbolic programming approach, wherein the data is matched to a set of predefined symbols.
We fit the trajectories of each keypoint in the trimmed time span to primitive shapes(circle, line, stationary)(Fig. \ref{fig_angles_prims}(b)). Note that the change detection is done on angles and lengths signals, whereas the primitives are fit on 2D coordinates of each keypoint. This symbolic representation brings us one step closer to the temporal semantics of the motion. Thus, they are better features for downstream tasks. Fig. \ref{fig_block_diagram} provides a block diagrammatic depiction of the process.


\begin{figure*}[ht]
\centering  
\includegraphics[height=8cm]{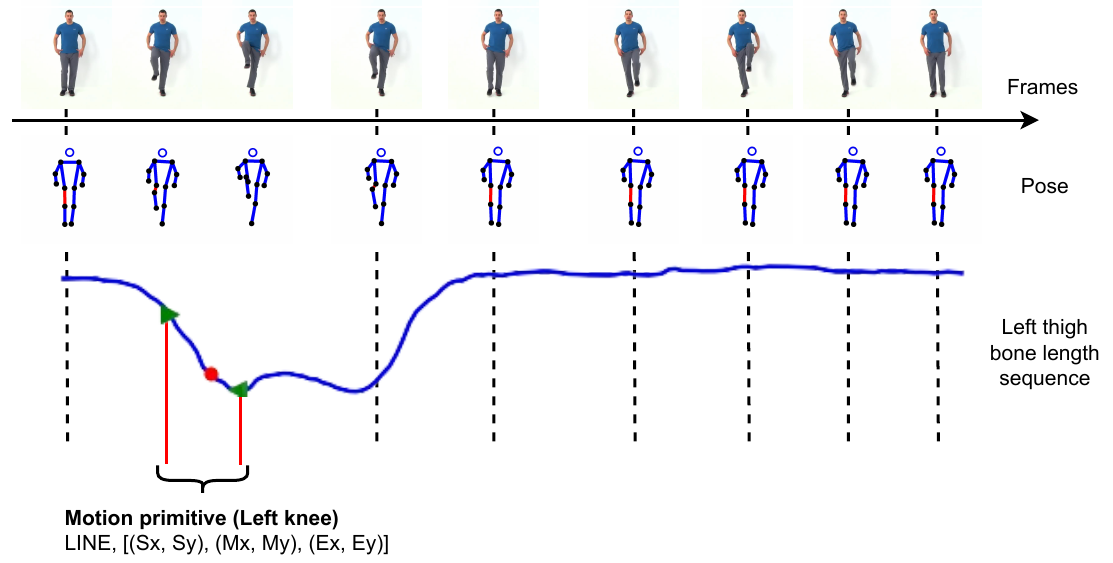}
\caption{The signal represents the variation of left thigh bone length(indicated in red) as a knee raise is performed. First, a meaningful segment is trimmed out using change detection, followed by fitting a primitive to each trajectory of the keypoints within that segment. In the figure, a representative fit to one of the keypoint(left knee) is indicated.}
\end{figure*}

\subsection{Temporal Segmentation using Change Detection}

Given an incoming time series, we want to find out the window of time around in which a meaningful change in body part motion has happened. We define the interval from the starting time of the change to the ending time of the change as the segment interval. In the proposed method, we perform change detection on joint angles and bone lengths, as they are indicative of sub-actions. It was demonstrated in \cite{Dittakavi2022} that the feature constituting joints and angles gives the best performance in a pose-based classification task. 

\subsubsection{CuSum Change Detection}

Change detection algorithms work by identifying a shift in certain statistical characteristics of the input signal. Change detection proves to be particularly effective in this application because of the fact that human motions are constrained and hence always vary within particular ranges. Here, we use the CuSum(Cumulative Sum) change detection algorithm. CuSum is a classical change detection method developed by E.S. Page\cite{Page1954}. It detects a shift in the mean of the signal. It is a sequential method and hence can be executed on an online time series.   

We explain the method using Fig. \ref{fig_change_det}.
The time series in the figure is the variation of left shoulder angle of a subject performing a jumping jack. As it can be observed, the sub-action of raising the hand is signified by the ascending section of the signal. Such similar variations in joint angles and bone lengths serve as indicators for human sub-actions. 
We leverage this property of human motion to segment out relevant sections to be further represented. To do this, we need to detect the starting time and ending time of the change. In the above example, we are interested in determining the starting point and ending point of the $S$-shaped segment. We do this using change detection. Let $t_{\text{start}}, t_{\text{change}}, t_{\text{end}}$ denote the start time of the change, the time at which change is detected, and the end time of the change, respectively. 

In CuSum change detection, given input time series, $x[t]$, the cumulative sum of positive and negative changes ($g_t^+$ and $g_t^-$) are calculated and compared to a threshold. When the threshold is crossed, it is flagged that a change is detected, and the cumulative sums are reset to 0. Since both positive and negative changes are tracked, it can detect both increasing and decreasing changes. We make use of the CuSum implementation of the detecta python package \cite{detecta}. The detailed steps are given in Algorithm \ref{algo_cusum}.

\begin{algorithm}
\caption{CuSum Change Detection}
\begin{algorithmic}[1]
    \For{ \textbf{each} $t$}
        \State $s[t] = x[t] - x[t-1]$
        \State $g^+[t] = \max(g^+[t-1] + s[t], 0)$
        \State $g^-[t] = \max(g^-[t-1] + s[t], 0)$

        \If{ $g^+[t] < 0$ }
            \State $g^+[t] = 0$
            \State $t^+=t$
        \EndIf

        \If{ $g^-[t] < 0$ }
            \State $g^-[t] = 0$
            \State $t^-=t$
        \EndIf
        
        \If{$g^+[t] > \text{threshold}$ or $g^-[t] > \text{threshold}$}
            \State $t_{\text{change}} = t$

            \If{ $g^+[t] > \text{threshold}$ }
                \State $t_{\text{start}} = t^+$
            \Else
                \State $t_{\text{start}}=t^-$
            \EndIf

            \State $g^+[t] = 0$
            \State $g^-[t] = 0$
            
        \EndIf
    \EndFor
\end{algorithmic}
\label{algo_cusum}
\end{algorithm}






\subsubsection{End-time of Change}

After using CuSum, we have the start point of change and the point of change, but we do not have the end point of change. We estimate the end of the change by fitting two piece-wise lines, where the second line is a constant. 
By fitting such a function, we are assuming that the change continues to happen for some more time, and then it reaches a constant value. We are interested in the time when the transition happens, and we define that as the end time of the change.

The following piece-wise linear function is fit to the time series between $t_{\text{change}}$ and $t_{\text{change}}$ + ($t_{\text{change}}$ - $t_{\text{start}}$). The choice of this interval for fitting is under the assumption that the rate of the range would remain approximately constant.

\[
y =
\begin{cases}
    ax + b & \text{when } x < c \\
    d & \text{when } x \geq c
\end{cases}
\]

$c$ is the transition point, and is taken as $t_{\text{end}}$.  

\begin{figure}[htbp]
\centering
\includegraphics[height=5cm]{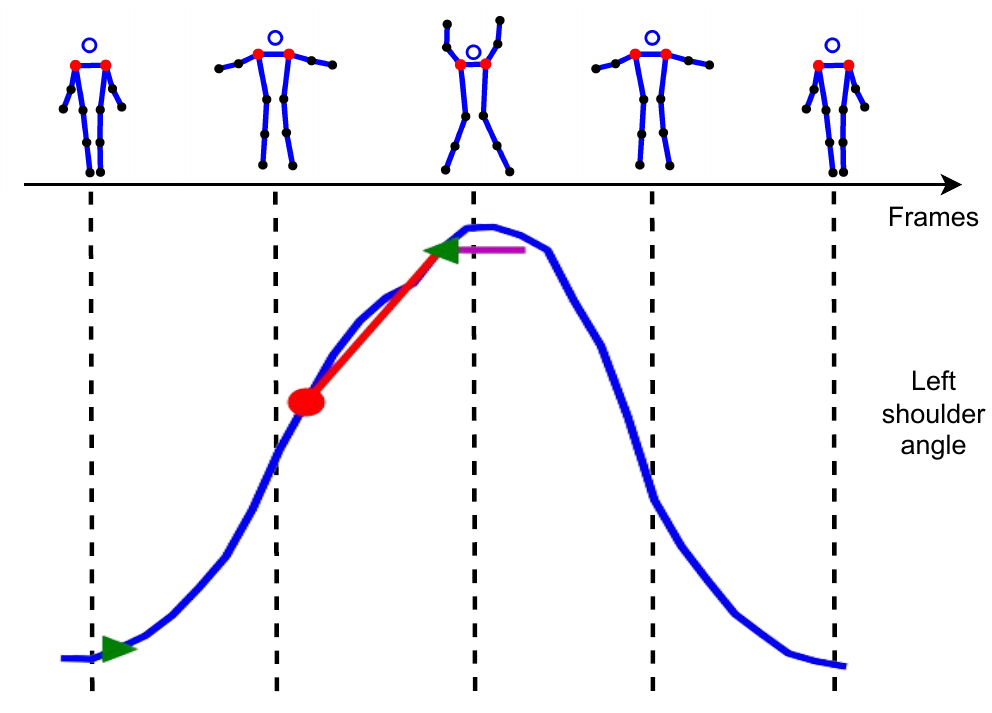} 
\caption{In a jumping jack instance, a change was detected in the left shoulder joint angle in the upward motion.}
\label{fig_change_det}
\end{figure}

In Fig. \ref{fig_change_det}, the red dot indicate $t_{\text{change}}$   and the green arrows indicate $t_{\text{start}}$ and $t_{\text{end}}$. The piece-wise linear fit for estimation of $t_{\text{end}}$ is also indicated in the figure. As it can be observed, the transition point of the two lines is $t_{\text{end}}$. 

Now we have start and end times. Action is segmented from $t_{\text{start}}$ to $t_{\text{end}}$. Segments obtained this way capture meaningful movements like arm swings, knee raises, etc.

\subsection{Fitting Motion Primitives}
Over the window of frames that we segmented out, each keypoint has a sequence of 2D coordinates. We fit a primitive shape like a line or circle to each keypoint sequence. This approach falls under symbolic programming, and its application to represent motion was introduced in \cite{Kulal2021}. In this manner, we obtain a higher-level representation of the sub-action.






We use least-squares to fit a given set of points to line or circle($(x_i - x_c)^2 + (y_i - y_c)^2 = r^2$). Further, polynomial fitting with respect to time is done in order to get a smoother parametric variation with time. 
x and y coordinates of a line are fit as : $\text{line}_x = a_x t^2 + b_x t + c_x$ and $\text{line}_y = a_y t^2 + b_y t + c_y$, where $t$ is the parametric time and $a_x,b_x,c_x,a_y,b_y,c_y$ are fitted parameters.
Similarly, for circle, a polynomial fit for angle($\theta$) is done. First $\theta$ value for each time value is found using $x = x_c + r \cos{\theta}$ and $y = y_c + r \sin{\theta}$. Then those $(t , \theta)$ value pairs are fitted as a polynomial: $\theta = a_c t^2 + b_c t + c_c$.

If the deviation of a set of coordinates from their mean is smaller than a threshold, those are fitted as a stationary primitive and represented using the mean point($\overline{x},\overline{y}$).

The sequence of coordinates for each keypoint is fitted to all three primitives using the aforementioned methods, and the one with the minimum error is chosen as the fit. 


\subsection{Action Representation}

Next, we compute the start, mid, and end coordinates of each keypoint of each primitive by using the fitted parameters. \textit{fit(t=0)}, \textit{fit(t=T/2)} and \textit{fit(t=T)} gives start, mid and end points respectively. Our final motion primitive representation is $[ \text{PRIM TYPE}, [(Sx, Sy), (Mx, My), (Ex, Ey)] ]$, where ``PRIM TYPE" is the fitted symbol. In addition, we determine the start, mid, and end values of each joint angle and each bone length signal. i.e. we have $\text{joint}_i$: start, mid, end and $\text{bone}_i$ : start, mid, end.  

These higher-level representations are very effective and can be used for a wide range of downstream tasks. It encapsulates the starting, middle, and ending points of an action sequence and also the approximate trajectory shape of each keypoint. We demonstrate the applicability of this representation in repetition detection. 

Using start, mid, and end values, we can figure out whether an angle is increasing, or a part is moving left or right, etc. In this way, they capture relevant information about the dynamics of that sub-action. Therefore, these serve as features of good discriminative power for recognition tasks as well.

\section{Repetition Detection}

We use this action representation for class agnostic repetition detection, which is one example of a downstream task. 

We propose a consecutive match search algorithm for repetition detection. Match is defined using an error metric between the start, mid, and end points of two primitives. The error is given by 
$|\overline{s}_i - \overline{s}_j| + |\overline{m}_i - \overline{m}_j| + |\overline{e}_i - \overline{e}_j|$, where $\overline{s}_i, \overline{m}_i,  \overline{e}_i, \overline{s}_j ,\overline{m}_j , \overline{e}_j$ are start, mid and end coordinates of two primitives $i$ and $j$ respectively of one keypoint, and the norm is L2 norm. This error is defined for all keypoints, and the total error is taken as the sum of all. Two motion primitives are said to match if the error is less than a threshold.

A buffer of the recent motion primitives is maintained. When a new motion primitive is encountered, it is matched with the existing primitives in the buffer. If a match is found, then consecutive matches are searched for by drawing more primitives and checking matches with adjacent primitives. This process is continued until a mismatch. A group of consecutive primitives that are matched belong to the same loop. The buffer is updated by removing the primitives which are matched. 
Counts get updated as the instances occur, as opposed to giving a total count of a given video offline.

\begin{figure}[htbp]
\centering
\includegraphics[width=\columnwidth]{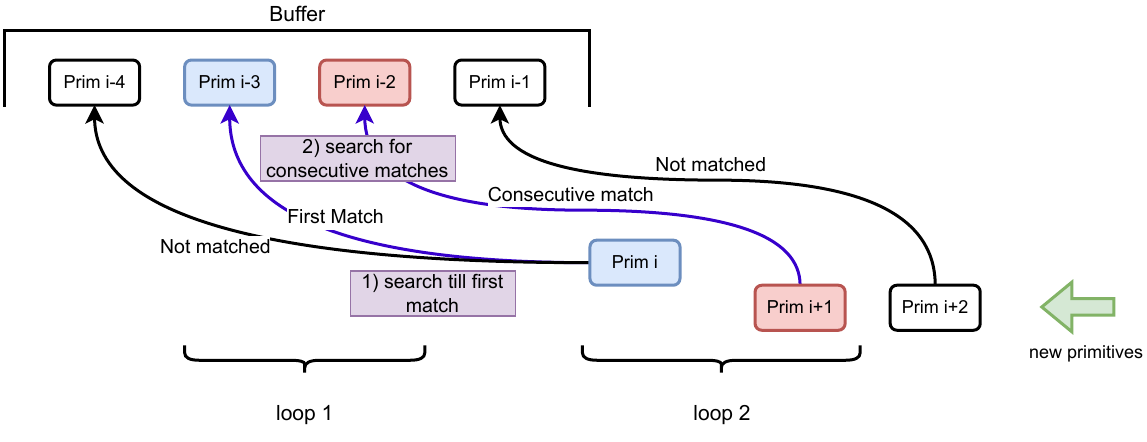} 
\caption{Once a match is found in the buffer, consecutive matches are searched for by drawing more primitives, and this continues until the match chain is broken. Groups of consecutive matches are loops.}
\label{fig_loop_det}
\end{figure}

\section{Experiments \& Results}

\begin{figure}[htbp]
\centering
\includegraphics[width=\columnwidth]{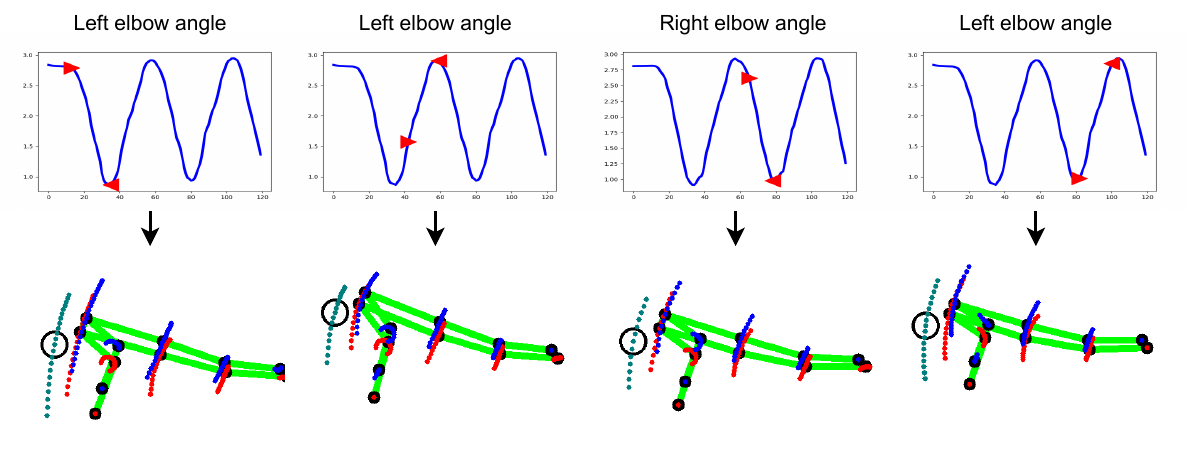} 
\caption{Change detection and primitive fitting on two instances of push-up}
\label{fig_pushup}
\end{figure}

\begin{figure}[htbp]
\centering
\includegraphics[width=5cm]{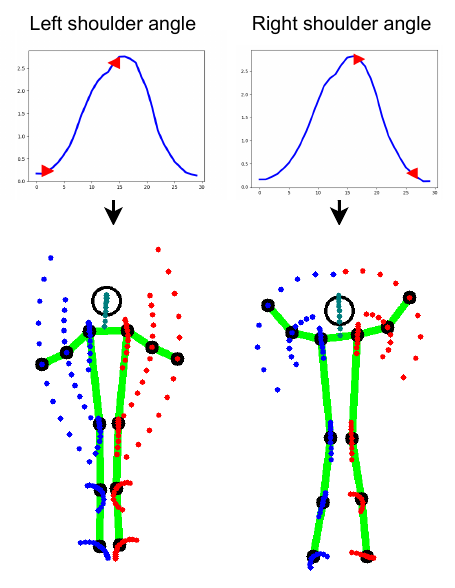} 
\caption{Change detection and primitive fitting on jumping jack}
\label{fig_jj}
\end{figure}

\begin{figure}[htbp]
\centering
\includegraphics[width=\columnwidth]{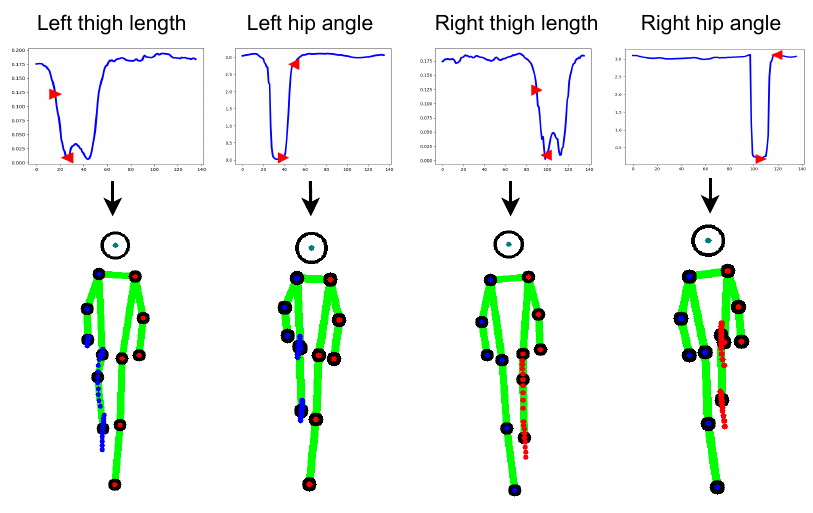} 
\caption{Change detection and primitive fitting on high knee raise}
\label{fig_hkm}
\end{figure}

\begin{figure}[htbp]
\centering
\includegraphics[width=\columnwidth]{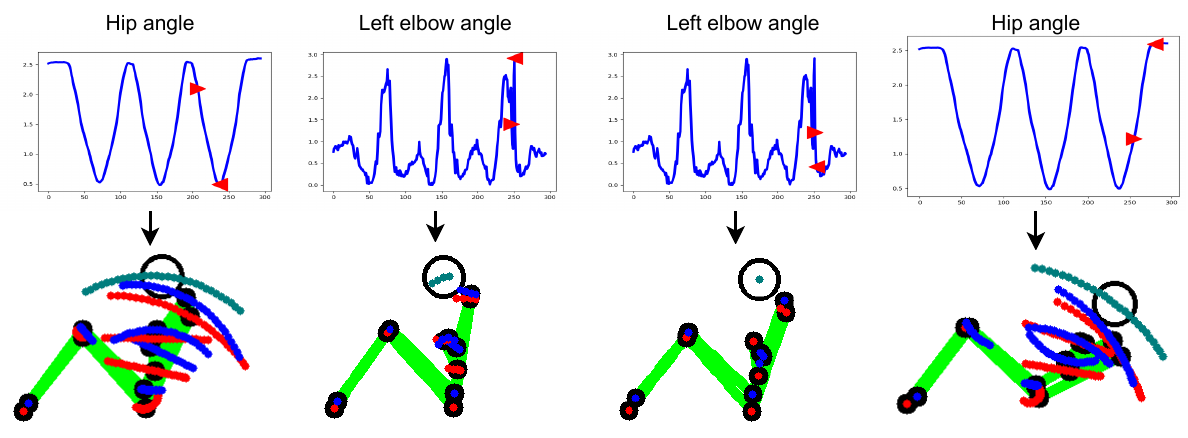} 
\caption{Change detection and primitive fitting on one instance of crunches (showing false positives) }
\label{fig_crunches}
\end{figure}

We demonstrate change detection and primitive fitting in 4 different exercise examples(Fig. \ref{fig_pushup}, Fig. \ref{fig_jj}, Fig. \ref{fig_hkm}, Fig. \ref{fig_crunches}). The segmentation window found using change detection is indicated by the red arrows in the angle/length sequences. Corresponding primitive fit(for each keypoint) is overlaid on the skeletons below. The trajectories that appear curvy are those which were fitted as circles, those which appear straight were fitted as lines and those which are single points were fitted as stationary.
It can be observed that majority of the extracted primitives are meaningful. Fig. \ref{fig_crunches} shows an example of false positives, where primitives are detected though not meaningful. From the figure, it can be seen that the middle two primitives are segmented based on change in elbow-angle signal, which is irrelevant in crunches. Ideally, we want two primitives to be detected(the up motion and the down motion), but two additional noisy primitives were added.

In \cite{Kulal2021}, a loop detection method that uses similar start, mid, and end points of extracted primitives is proposed. However, those primitives are extracted offline from videos. We compare our method with that method in terms of per-class percentage accuracy on repetition counting. Percentage accuracy is defined as $( 1-\frac{ \left| \text{Predicted count}-\text{True count} \right| }{\text{True count}} * 100 )$.
The validation is done on moticon dataset from \cite{Kulal2022} after manually labeling the number of counts of actions in the videos. We validate the methods only on videos where the count is more than 1, as the algorithms are based on finding matches and hence require at least two instances. Table \ref{tab1} shows the per-class percentage accuracy of our online method and the aforementioned method. 
We managed to get either comparable or better results in each class. Note that this was achieved by operating online, which is the key contribution. The offline method failed to do prediction on bulk of arm-cross-swings instances, and hence it was left out. Meanwhile, the proposed method gives consistent results for all classes. 

\begin{table}[htbp]
\caption{loop counting on moticon dataset}
\begin{center}
\begin{tabular}{|c|c|c|c|}
\hline
\textbf{\textit{Class}}& \textbf{\textit{\# vid}}& \textbf{\textit{Online \% acc.}}& \textbf{\textit{Offline \% acc.}} \\
\hline
Jumping jack& 89& 73.9\%& 78.4\% \\
High knee march& 15& 83.7\%& 74.7\% \\
Arm cross swings& 39& 72.0\%& -  \\
\hline
All classes& 136& \textbf{74.0\%}& 78.1\%  \\
\hline
\end{tabular}
\label{tab1}
\end{center}
\end{table}

Further, we collected more exercise videos and evaluated the method using those. The data collection involved 3 subjects, each performing 4 exercises(push-ups, crunches, arm raises, arm rotations). Each participant performed 5 instances of varying counts for each exercise, totaling 60 videos of different counts for different exercises. We annotated each video with the actual count and then calculated the percentage accuracy of the proposed method on this data. The exercises were performed in different view orientations w.r.t the camera in order to check the performance of the method in such cases. Table \ref{tab2} shows per class percentage accuracy on repetition counting using the proposed method. 
It can be observed that the accuracy is high for simple motion exercises like arm raises and push-ups, where the movements remain mostly regular. There is a higher chance of irregularity in the repeated motion for exercises like crunches and arm rotation. This factor has an effect on the performance because the method relies on similarity matching.

\begin{table}[htbp]
\caption{Loop counting on custom data}
\begin{center}
\begin{tabular}{|c|c|c|}
\hline
\textbf{\textit{Class}} & \textbf{\textit{\# videos}} & \textbf{\textit{Percentage accuracy}} \\
\hline
Push ups& 15& 81.5\%  \\
Crunches& 15& 58.3\%  \\
Arm raises& 15& 100\%  \\
Arm rotations& 15& 61.8\%  \\
\hline
All classes& 60& \textbf{76.0\%} \\
\hline
\end{tabular}
\label{tab2}
\end{center}
\end{table}

\section{Discussion}
The main advantage of the proposed method is that it operates online. Furthermore, the choice of symbolic representation makes it compositional, generalizable, and interpretable. The compositionality here is with respect to body parts. A human activity can be composed of elementary actions of each part. It can be argued that these elementary primitives are finite in number as human motion is constrained. Then, any activity can be represented as a composition of those finite primitives. This makes it generalizable. 
A part-based representation also proves to be advantageous in scenarios where the view is incomplete, such as when the camera is positioned such that just a portion of the body is visible. Since we explicitly know which symbol a part is fit to, the representation is interpretable. Also, this method doesn't require any data for training and instead relies on domain knowledge for good choices for primitives. 

Noisy involvement of other body parts when performing an exercise will compromise the effectiveness of the method. For e.g., if a person is randomly moving his arms while performing a knee raise, those hand motions might get triggered as changes. In that case, the segmentation done is not meaningful (Fig. \ref{fig_crunches}).
One way to address this drawback is to use a consensus of change detections over multiple time series instead of considering only a single series. 
Usage of adaptive thresholds for different signals may also be a promising approach.

We use bone lengths along with joint angles because we are working with 2D coordinates. We expect bone lengths to capture surrogates of some of the 3D motions. If we use a 3D pose estimation method directly, then we need only use angle sequences for change detection, and further we can directly fit 3D curves as motion primitives. However, most of the 3D pose methods are noisy along the depth dimension.

\section{Conclusion}
We address the problem of online action representation. We use a change detection algorithm, which has been shown to give meaningful segments for further symbolic fitting. Using a symbolic programming approach allows us to use this method in sparse data scenarios. Additionally, the nature of segmentation and primitive fitting captures the compositional aspect of actions as a consequence. This representation can effectively work for many downstream tasks, and more importantly, it can be operated online, which is the primary gap that is addressed.


\bibliographystyle{IEEEtran}
\bibliography{references}

\begin{thebibliography}{10}
\providecommand{\url}[1]{#1}
\csname url@samestyle\endcsname
\providecommand{\newblock}{\relax}
\providecommand{\bibinfo}[2]{#2}
\providecommand{\BIBentrySTDinterwordspacing}{\spaceskip=0pt\relax}
\providecommand{\BIBentryALTinterwordstretchfactor}{4}
\providecommand{\BIBentryALTinterwordspacing}{\spaceskip=\fontdimen2\font plus
\BIBentryALTinterwordstretchfactor\fontdimen3\font minus \fontdimen4\font\relax}
\providecommand{\BIBforeignlanguage}[2]{{%
\expandafter\ifx\csname l@#1\endcsname\relax
\typeout{** WARNING: IEEEtran.bst: No hyphenation pattern has been}%
\typeout{** loaded for the language `#1'. Using the pattern for}%
\typeout{** the default language instead.}%
\else
\language=\csname l@#1\endcsname
\fi
#2}}
\providecommand{\BIBdecl}{\relax}
\BIBdecl

\bibitem{Sabater2021}
A.~Sabater, L.~Santos, J.~Santos-Victor, A.~Bernardino, L.~Montesano, and A.~C. Murillo, ``One-shot action recognition in challenging therapy scenarios,'' in \emph{2021 IEEE/CVF Conference on Computer Vision and Pattern Recognition Workshops (CVPRW)}.\hskip 1em plus 0.5em minus 0.4em\relax Los Alamitos, CA, USA: IEEE Computer Society, jun 2021, pp. 2771--2779.

\bibitem{Zhao2023}
Z.~Zhao, S.~Kiciroglu, H.~Vinzant, Y.~Cheng, I.~Katircioglu, M.~Salzmann, and P.~Fua, ``3d pose based feedback for physical exercises,'' in \emph{Computer Vision -- ACCV 2022}, L.~Wang, J.~Gall, T.-J. Chin, I.~Sato, and R.~Chellappa, Eds.\hskip 1em plus 0.5em minus 0.4em\relax Cham: Springer Nature Switzerland, 2023, pp. 189--205.

\bibitem{Fieraru2021}
M.~Fieraru, M.~Zanfir, S.~C. Pirlea, V.~Olaru, and C.~Sminchisescu, ``Aifit: Automatic 3d human-interpretable feedback models for fitness training,'' in \emph{2021 IEEE/CVF Conference on Computer Vision and Pattern Recognition (CVPR)}, 2021, pp. 9914--9923.

\bibitem{Dittakavi2022}
B.~Dittakavi, D.~Bavikadi, S.~V. Desai, S.~Chakraborty, N.~Reddy, V.~N. Balasubramanian, B.~Callepalli, and A.~Sharma, ``Pose tutor: An explainable system for pose correction in the wild,'' in \emph{2022 IEEE/CVF Conference on Computer Vision and Pattern Recognition Workshops (CVPRW)}, 2022, pp. 3539--3548.

\bibitem{Kulal2021}
S.~Kulal, J.~Mao, A.~Aiken, and J.~Wu, ``Hierarchical motion understanding via motion programs,'' \emph{2021 IEEE/CVF Conference on Computer Vision and Pattern Recognition (CVPR)}, pp. 6564--6572, 2021.

\bibitem{Levy2015}
O.~Levy and L.~Wolf, ``Live repetition counting,'' in \emph{2015 IEEE International Conference on Computer Vision (ICCV)}, 2015, pp. 3020--3028.

\bibitem{Dwibedi2020}
D.~Dwibedi, Y.~Aytar, J.~Tompson, P.~Sermanet, and A.~Zisserman, ``Counting out time: Class agnostic video repetition counting in the wild,'' \emph{2020 IEEE/CVF Conference on Computer Vision and Pattern Recognition (CVPR)}, pp. 10\,384--10\,393, 2020.

\bibitem{mediapipe}
C.~Lugaresi, J.~Tang, H.~Nash, C.~McClanahan, E.~Uboweja, M.~Hays, F.~Zhang, C.-L. Chang, M.~G. Yong, J.~Lee, W.-T. Chang, W.~Hua, M.~Georg, and M.~Grundmann, ``Mediapipe: A framework for building perception pipelines,'' \emph{ArXiv}, vol. abs/1906.08172, 2019.

\bibitem{Page1954}
E.~S. PAGE, ``{CONTINUOUS INSPECTION SCHEMES},'' \emph{Biometrika}, vol.~41, no. 1-2, pp. 100--115, 06 1954.

\bibitem{detecta}
M.~Duarte, ``detecta: A python module to detect events in data,'' \url{https://github.com/demotu/detecta}, 2020.

\bibitem{Kulal2022}
S.~Kulal, J.~Mao, A.~Aiken, and J.~Wu, ``Programmatic concept learning for human motion description and synthesis,'' in \emph{2022 IEEE/CVF Conference on Computer Vision and Pattern Recognition (CVPR)}.\hskip 1em plus 0.5em minus 0.4em\relax Los Alamitos, CA, USA: IEEE Computer Society, jun 2022, pp. 13\,833--13\,842.

\end{thebibliography}





\end{document}